\documentclass{article}
\usepackage{spconf,amsmath,epsfig}
\usepackage{placeins}
\usepackage{amsmath}
\usepackage{amssymb}  
\usepackage{amsfonts} 
\usepackage{bbm}
\usepackage{pifont}   
\usepackage{threeparttable}
\usepackage{todonotes}
\usepackage{comment}
\providecommand{\doi}[1]{doi: {\footnotesize \href{http://dx.doi.org/#1}{\path{#1}}}}

\usepackage[pdftex=true,breaklinks=true,hidelinks=true,colorlinks=true,citecolor=blue]{hyperref}

\usepackage[square,numbers]{natbib}

\title{Lightweight CNNs for Embedded SAR Ship Target
Detection and Classification}
\name{%
  Fabian~Kresse\textsuperscript{1,3}\thanks{%
 F.~Kresse and G.~Pilikos performed this work while at ESA and are no longer affiliated with the agency. \\
    Correspondence: \texttt{Nicolas.Floury@esa.int}%
  },
  Georgios~Pilikos\textsuperscript{1,4},
  Mario~Azcueta\textsuperscript{2},
  Nicolas~Floury\textsuperscript{1}%
}

\address{%
  \parbox{0.9\textwidth}{%
    \footnotesize
    \textsuperscript{1}Wave Interaction \& Propagation, RF Payloads \& Tech.\ Div., Elec.\ Dept., ESA/ESTEC, NL-2200AG Noordwijk, The Netherlands\\
    \textsuperscript{2}Copernicus Sentinel-1 Payload, Copernicus Space Component, ESA/ESTEC, NL-2200AG Noordwijk, The Netherlands\\
    \textsuperscript{3}Institute of Science and Technology Austria (ISTA), AT-3400 Klosterneuburg, Austria \\
    \textsuperscript{4}Department of Neuroscience, Erasmus MC, NL-3000CA Rotterdam, The Netherlands
  }
}
\begin{document}

\maketitle

\begin{abstract}
Synthetic Aperture Radar (SAR) data enables large-scale surveillance of maritime vessels. However, near-real-time monitoring is currently constrained by the need to downlink all raw data, perform image focusing, and subsequently analyze it on the ground. On-board processing to generate higher-level products could reduce the data volume that needs to be downlinked, alleviating bandwidth constraints and minimizing latency. However, traditional image focusing and processing algorithms face challenges due to the satellite's limited memory, processing power, and computational resources. This work proposes and evaluates neural networks designed for real-time inference on unfocused SAR data acquired in Stripmap and Interferometric Wide (IW) modes captured with Sentinel-1. Our results demonstrate the feasibility of using one of our models for on-board processing and deployment on an FPGA. Additionally, by investigating a binary classification task between ships and windmills, we demonstrate that target classification is possible.
\end{abstract}
\begin{keywords}
Deep Learning, raw echo data, ship detection, synthetic aperture radar (SAR), Field-Programmable-Gate Array (FPGA)
\end{keywords}

\section{Introduction}

Synthetic Aperture Radar (SAR) satellite data enables all-weather maritime monitoring. Traditional, on-ground Constant False Alarm Rate (CFAR)~\cite{crisp2004state} detection on focused SAR images entails a costly downlink, focus, analyze pipeline, incurring latency and limited contact windows. Processing the data directly on the satellite significantly reduces the volume of data that needs to be downlinked by generating higher abstraction level outputs (e.g., pixel coordinates of detected ships) instead of raw data. Deep learning models have shown promising results in this context~\cite{li2023survey,pilikos2022raw, pilikos2024raw, ghiglione2024iac, zhang2020shipdenet}, offering the opportunity to optimize for inference on embedded devices straightforwardly. Yet, obtaining fully focused SAR images on satellite is computationally and memory intensive, prompting research into onboard ship detection using the intermediate raw~\cite{de2024ship, leng2023ship} or range-compressed data products~\cite{zeng2024incept, leng2022ship, joshi2019range}. Overall, an effective onboard SAR ship detection algorithm must process small data segments to accommodate limited on-board memory, maintain a compact model suitable for Field-Programmable-gate Arrays (FPGAs), directly output ship coordinates to reduce downlink and storage requirements, and still achieve high accuracy. While prior work addresses individual elements of this pipeline, none achieve real-time, accurate detection on unfocused or range-compressed data under embedded FPGA constraints.
\begin{figure*}[t!]
    \centering
    \includegraphics[width=1.0\linewidth]{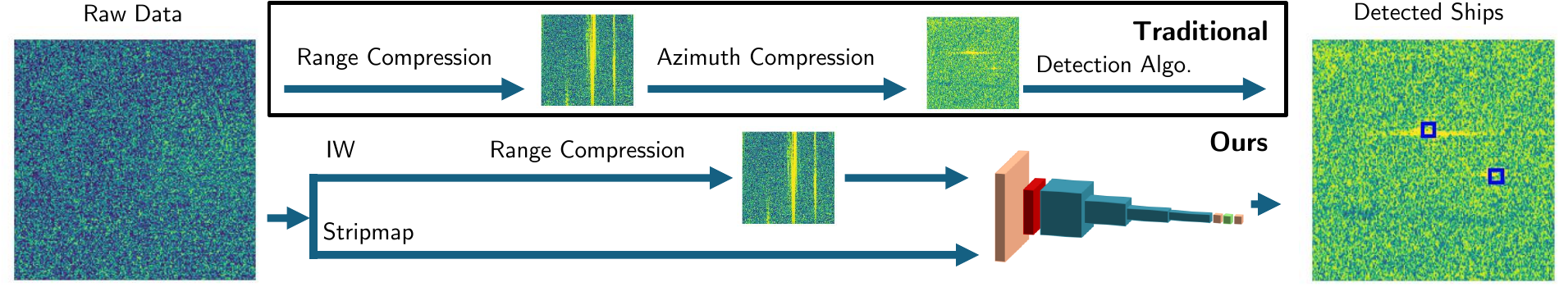}
    \caption{The traditional pipeline (black box) focuses data before detection. Our approach bypasses this: we detect directly on raw Stripmap echoes~\cite{de2024ship} and on range-compressed IW data.}
    \label{fig:pipeline}
\end{figure*}

In this work, we propose a range of lightweight deep-learning models designed for Stripmap and Interferometric Wide (IW) SAR data obtained with Sentinel-1, addressing the requirements for on-board data processing. Our model configurations, processing raw and range-compressed data, are one-stage detectors built on ResNet blocks~\cite{he2016deep}, allowing for flexible model sizing. They employ a single-stage detection layer as the final step, which performs coordinate predictions and target classification following the grid-based approach of the YOLO architecture~\cite{redmon2016you}. Our smallest model operating on Stripmap mode data delivers near-perfect ship detection results in our evaluation scenes. Additionally, we demonstrate that this model can be successfully deployed on a Xilinx Zynq UltraScale+ MPSoC ZCU104 FPGA, achieving sufficient throughput for real-time processing. For IW data, we evaluate the performance of multiple model sizes, achieving competitive results in open water scenes and offering valuable insights for future research. We also show that our model can perform binary target classification, distinguishing between windmills and ships in the IW dataset we employ.

\section{Proposed Deep Learning Algorithm}
\label{sec:models}

Fig.~\ref{fig:pipeline} shows the processing pipeline for both the traditional method, including range and azimuth compression, and the approach adopted by us. Similar to previous work \cite{de2024ship}, for Stripmap mode, we operate directly on raw data, while for IW data, we perform range-compression.

\textbf{Stripmap Preprocessing:} We shift the raw data by half the chirp length in the range dimension so that the feature response aligns with the Single-Look Complex (SLC) labels, accounting for mode- and chirp-specific acquisition offsets.

\textbf{IW Preprocessing:} Range-compressed IW features and SLC labels are misaligned, so we apply a locally consistent mapping between their pixel spaces. We compute the offset from the center pixel of each range-compressed crop and use it to shift the corresponding label crop.

\textbf{Model: } We employ a YOLO-style  architecture for its efficient single-stage design, enabling fast inference without the overhead of region proposals~\cite{redmon2016you}. The backbone consists of four layers of ResNet blocks, preceded by a $7\times7$ convolution with 64 kernels. As in previous work, we treat the complex-valued SAR data as two separate input channels~\cite{de2024ship}. We apply the network to crops of the original SAR image. The network outputs predictions on a YOLO-style grid, where each grid cell predicts the coordinates $(x, y)$ and an objectness score indicating the presence of a target. Since ships in our datasets occupy a narrow range of sizes, we omit multi-scale anchor-based detection typical in YOLO. We also perform binary classification between ships and windmills for IW data, adding two additional outputs. Our output grid has size $S\times S$, with each cell corresponding to a $32 \times 32$ pixel region in the input data. For example, an input crop size of $128\times128$ results in an output grid of size $4\times4$. The model configurations evaluated in our experiments are listed in Table~\ref{table: model configs}.

\begin{table}[h!]
    \centering
    \caption{Each Resnet-block contains two convolutions with kernels of size $3\times3$, Batchnorm and ReLU activation.}
    \label{table: model configs}
    \begin{tabular}{c|c|c}
         \textbf{Param. (Size)} & \textbf{Blocks per Layer} & \textbf{Channels} \\
         \hline  
          96800 (S) & 1, 2, 2, 1 & 16, 16, 32, 32\\
          1222368 (M) & 3, 4, 6, 3 & 64, 64, 64, 64\\
          11222880 (L) & 2, 2, 2, 2 & 64, 128, 256, 512\\
    \end{tabular}
    
\end{table}

\textbf{Loss Function:} We adopt the standard YOLO loss, omitting only the bounding‐box size regression term while retaining all other components unchanged.

\begin{figure*}[t!]
    \centering
    \begin{minipage}[b]{0.49\linewidth}
        \centering
        \includegraphics[width=\linewidth]{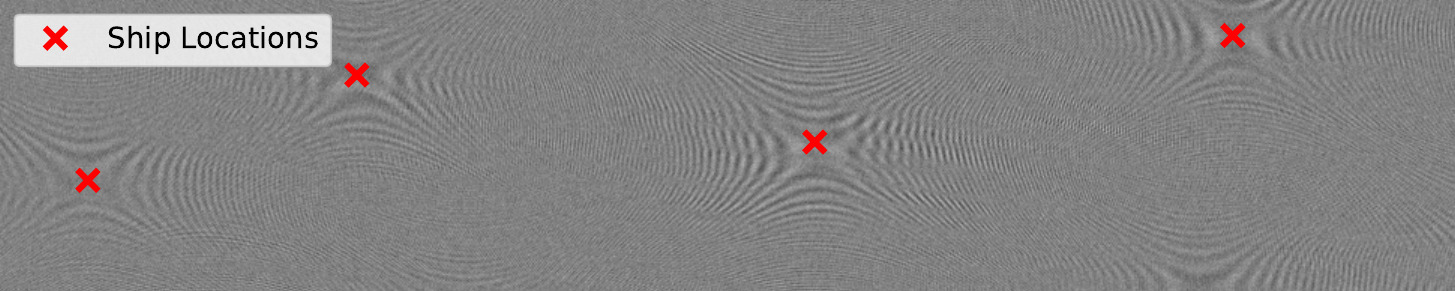}
        \label{fig:image1_s}
    \end{minipage}
    \hfill
    \begin{minipage}[b]{0.49\linewidth}
        \centering
        \includegraphics[width=\linewidth]{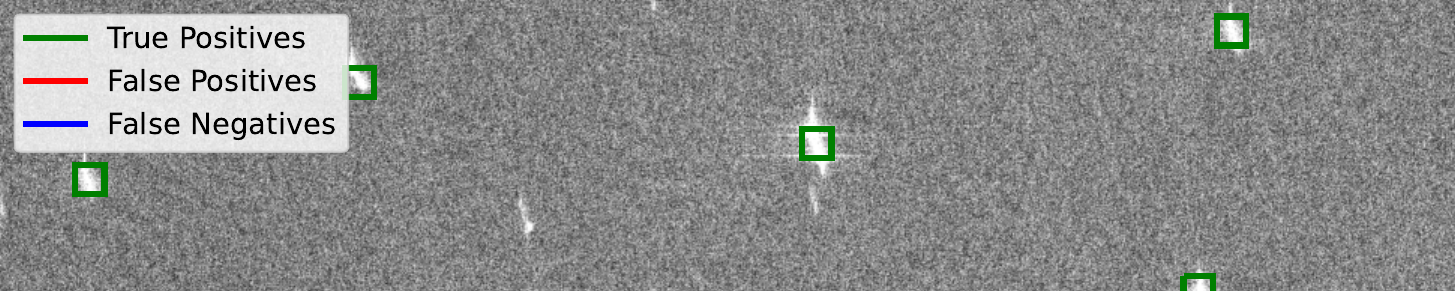}
        \label{fig:image2_s}
    \end{minipage}
    \caption{Left: real part of raw Stripmap SAR data, shifted by half-chirp to align SLC labels; 
Right: manually labeled SLC image overlaid with the network’s post-processed detections (contains modified Copernicus Sentinel Data~\cite{copernicus_data_space}).
}

    \label{fig:Stripmap quality}
\end{figure*}

\textbf{Prediction of Ship Locations:} After obtaining predictions, we assign a fixed 50-pixel bounding box to each detected ship. As in the YOLO pipeline~\cite{redmon2016you}, we apply non-maximum suppression (NMS) to remove overlapping boxes, keeping the one with the highest confidence. The acceptance threshold is set as the lower of the Youden J threshold~\cite{youden1950index} and the minimum distance threshold from validation data, rounded down to two decimals. We compute a distance-based \(F1\) score \(F1_{30}\) counting a prediction as correct if it lies within 30 pixels of a ground-truth label; unmatched predictions and labels are false positives and negatives, respectively. Given SAR resolutions (5 m × 5 m for Stripmap, 5 m × 20 m for IW), this tolerance equals 150 m in range and up to 600 m in azimuth, accommodating minor localization errors while remaining well below the fixed box size.

\section{Experiments}
\label{sec:experiments}
\begin{figure*}[t!]
    \centering
    \begin{minipage}[b]{0.49\linewidth}
        \centering
        \includegraphics[width=\linewidth]{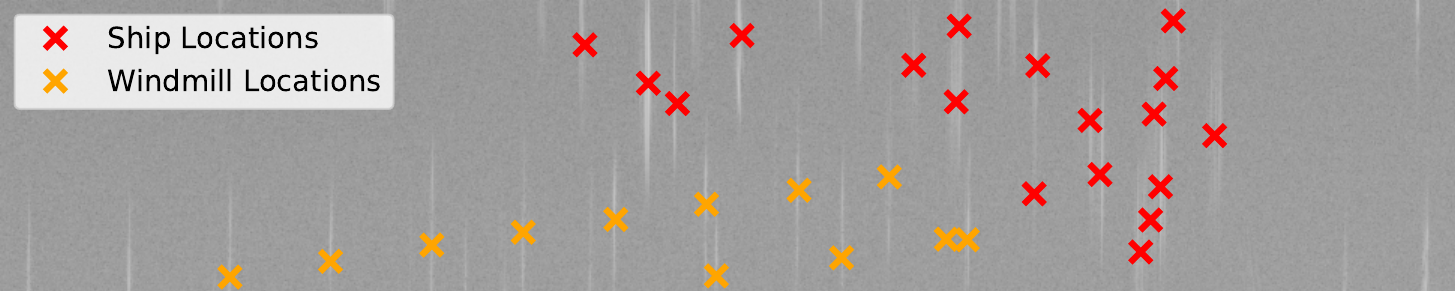}

        \label{fig:image1}
    \end{minipage}
    \hfill
    \begin{minipage}[b]{0.49\linewidth}
        \centering
        \includegraphics[width=\linewidth]{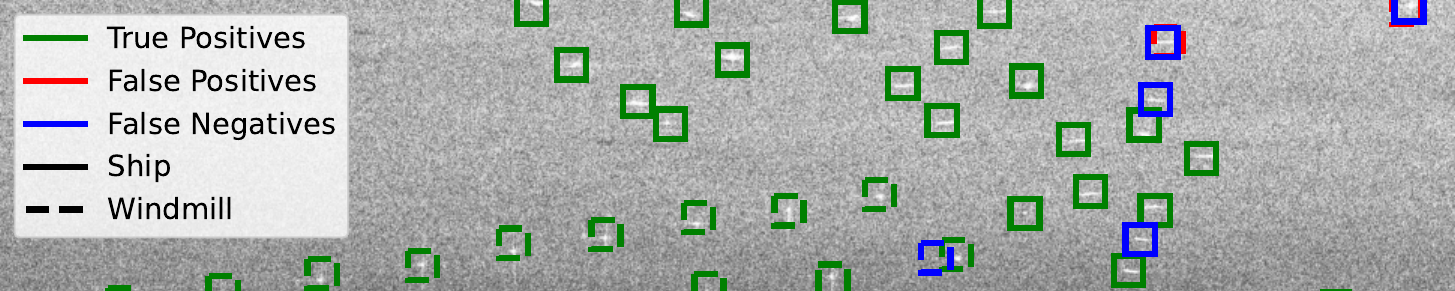}

    \end{minipage}
    \caption{Left: range-compressed IW SAR crop (input to the network) with center-based offset correction (Sec.~\ref{sec:models}); 
Right: post-processed detections overlaid on the labeled SLC image for the (M) model (contains modified Copernicus Sentinel Data~\cite{copernicus_data_space}).
}
        \label{fig:image2}
\end{figure*}

We evaluate our proposed model using two different datasets. The first dataset consists of raw SAR Sentinel-1 Stripmap mode (S6) VV polarization data, as previously utilized by~\cite{de2024ship}. We show that the model trained on this dataset can be deployed on an FPGA, achieving real-time throughput.  We then investigate an IW dataset from the Shanghai port. For this dataset, we show good off-shore detection performance and the ability of our model to distinguish between windmills and ships. Both our datasets where originally obtained from the Copernicus Data Space Ecosystem~\cite{copernicus_data_space}.  All experiments with standard deviations given were performed with three seeds.

\subsection{Sentinel-1 Stripmap Mode}

The dataset from~\cite{de2024ship}, consist of a total of 12 SAR images from the São Paulo port, with one image used for validation (84 ships), two for testing (155 ships), and the remaining for training (726 ships). We investigate the performance of the smallest model outlined in Section~\ref{sec:models} with a crop size of $128 \times 128$, achieving an $F1_{30}$ score of $0.98 \pm 0.00$ and $0.97 \pm 0.01$ on our first and second test image (see Fig.~\ref{fig:Stripmap quality} for qualitative results). After manually inspecting the few erroneous predictions, these can be attributed to double detections of ships, the NMS not being aggressive enough, and label ambiguities.

\subsection{Embedded Inference for Sentinel-1 Stripmap Mode}

We deployed our 8-bit AdaQuant-quantized~\cite{hubara2020improving} model (via Vitis AI 3.0~\cite{vitisAI}) on a Zynq UltraScale+ MPSoC ZCU104 FPGA with no accuracy loss. Real-time operation requires $\geq 2027$ FPS (PRF = 1664 lines/s $\times$ 19950 samples/line); DPU inference (excluding preprocessing \& NMS), achieved $3527 \pm 23$ FPS with four CPU threads, exceeding the target.

\subsection{Sentinel-1 Interferometric Wide Mode}

Our hand-labeled IW dataset from the Shanghai port comprises 10 bursts of size $20760\times1617$ at $5\times20$ m resolution. We split 8/1/1 bursts for training/validation/test sets. The training set contains 1163 ships and 460 windmills; after excluding near-shore ships, the test set contains 66 ships and 19 windmills. Preliminary experiments with ships located close to shore showed significant performance degradation, possibly due to the complex and ambiguous backscatter in these areas. Therefore, we exclude them from our final evaluation.

The IW data presents additional challenges due to the continuous antenna pattern steering in azimuth during the acquisition and the more complex nature of the scenes investigated. Initial attempts using our models with small input sizes on raw IW data did not yield adequate performance. As a result, we utilized range-compressed data.

Table~\ref{table:models iw} reports detection metrics and class-wise $F1_{30}$ (see Fig.~\ref{fig:image2} for quantitative results); class scores exclude the other class’s labels to isolate per-class performance. Our smallest model already performs well, and increasing crop size and parameter count further boosts $F1_{30}$, though gains plateau—likely due to overfitting. After manual inspection, a large number of remaining errors arise from closely spaced ships, due to NMS limitations, and ambiguous labels.

\begin{table}[h!]
    \caption{Comparison of Models: Results on Interferometric Wide data for off-shore ships on our test image (Range Compressed Input Data). Input denotes the crop size.}
    \centering
    \begin{threeparttable}
    \begin{tabular}{c|c|c|c|c}
    Size & Input  & $F1_{30}$ & Ship $F1_{30}$ & Wind. $F1_{30}$ \\
    \hline
    L  & $256$ & $0.87 \pm 0.01$ & $0.77 \pm 0.01$& $0.71 \pm 0.04$\\
    M & $256$ & $0.87 \pm 0.05$ & $0.78 \pm 0.05$& $0.78 \pm 0.02$\\
    S  & $256$ & $0.79 \pm 0.01$ & $0.56 \pm 0.07$& $0.54 \pm 0.06$\\
    S  & $128$ & $0.72 \pm 0.04 $ & $0.52 \pm 0.02$& $0.51 \pm 0.03$\\
    \end{tabular}
    \begin{tablenotes}
        \item
    \end{tablenotes}
    \end{threeparttable}
    \label{table:models iw}
\end{table}

\section{Discussion}
\label{sec:discussion}
We conduct experiments on raw Stripmap and range-compressed IW data. For Stripmap, excellent results are achieved by shifting the raw input by half the chirp size, bypassing the computationally expensive image-focusing step and enabling direct predictions on small crops of raw SAR data. The model remains compact in both parameters and forward-pass complexity, and we deploy it on a Zynq UltraScale+ MPSoC ZCU104 FPGA, demonstrating suitability for real-time embedded processing. However, the Stripmap scenes used here are relatively simple with similar backscatter, so future work should test robustness under more complex sea conditions and diverse environments.

In our evaluation of IW data, we find that raw data alone, in contrast to Stripmap data, does not yield satisfactory results. As a result, we perform range-compression, resulting in improved performance. We attribute this improvement to the target energy being more concentrated and, hence being easier to identify with the small Field-of-View of our neural network. By increasing both the input crop size and the network complexity compared to our Stripmap model, we achieve good performance on IW data. Additionally, we demonstrate successful target classification, performing binary classification between ships and windmills. 

A main limitation of our method is that it struggles to detect ships very close to shore, so we exclude these cases from the final evaluation. This low performance is likely due to strong backscatter from surrounding structures and vessels, and may be mitigated with more diverse training data.

Both the limitations observed in Stripmap and IW data can be largely attributed to the availability and quality of the datasets. A key challenge is the lack of sufficiently large and diverse labeled raw SAR datasets, particularly with varying sea conditions. The upcoming Sentinel-1 satellites, equipped with AIS tracking antennas, holds promise for addressing this gap by potentially enabling automated labeling through AIS data correlation. Additionally, since numerous algorithms already exist for processing SLC images, future work could leverage these methods to generate large-scale datasets by aligning SLC-based detections with raw SAR data, further expanding the pool of labeled data for training and evaluation.
\section{Contributions}
\label{sec:conclusion}

We demonstrated the feasibility of real-time ship detection using deep-learning models applied to raw Stripmap data. Furthermore, we deploy our Stripmap model on a Zynq UltraScale+ MPSoC ZCU104 FPGA, demonstrating its practical use for real-time, onboard processing in resource-constrained environments. For IW data, we show that using range-compressed data and larger input sizes improves detection and classification performance, enabling binary classification of ships and windmills. 
\paragraph*{Acknowledgment.}
The authors thank Sanath Muret, Max Ghiglione and Maris Tali for FPGA support and discussions.

\bibliographystyle{plainnat}
\bibliography{template-tex}

\small

\end{document}